\documentclass{article}

\usepackage{amsmath,graphicx}
\usepackage{amsfonts}
\usepackage{bm}
\usepackage{color}
\usepackage{booktabs}
\usepackage{array}
\usepackage[caption=false, font=footnotesize]{subfig}
\usepackage{url}
\usepackage{textcomp}
\usepackage{multirow}
\usepackage{cite}
\usepackage{authblk}

\title{Stochastic Gradient Descent for Spectral Embedding with Implicit Orthogonality Constraint}

\author[1]{Mireille El Gheche}
\author[2]{Giovanni Chierchia}
\author[1]{Pascal Frossard}
\affil[1]{Ecole Polytechnique F\'ed\'erale de Lausanne (EPFL), Signal Processing Laboratory (LTS4), Lausanne, Switzerland}
\affil[2]{Universit\'e Paris-Est, LIGM UMR 8049, CNRS, ESIEE Paris, Noisy-le-Grand, France}

\begin{document}
%
\maketitle

\begin{abstract}
In this paper, we propose a scalable algorithm for spectral embedding. The latter is a standard tool for graph clustering. However, its computational bottleneck is the eigendecomposition of the graph Laplacian matrix, which prevents its application to large-scale graphs. Our contribution consists of reformulating spectral embedding so that it can be solved via stochastic optimization. The idea is to replace the orthogonality constraint with an orthogonalization matrix injected directly into the criterion. As the gradient can be computed through a Cholesky factorization, our reformulation allows us to develop an efficient algorithm based on mini-batch gradient descent. Experimental results, both on synthetic and real data, confirm the efficiency of the proposed method in term of execution speed with respect to similar existing techniques.
\end{abstract}


\section{Introduction}
\label{sec:intro}

Graphs are appealing mathematical tools for modelling pairwise relationships between data points, without being limited by the rigid structure of Euclidean spaces. In a graph, the data points are represented as vertices, whereas the pairwise relationships between vertices are described by weighted edges. Examples of graph-based representations can be found in numerous application domains, such as biology, social networks, financial and banking, mobility and traffic patterns.

The main focus of this paper is spectral embedding, which aims at representing the nodes of a graph into a low-dimensional space that preserves the connectivity patterns described by the graph edges. In this regard, one of most popular approaches consists of embedding the graph vertices into a subspace spanned by the eigenvectors of the graph Laplacian matrix corresponding to the $K$ smallest eigenvalues \cite{Schaeffer2007, Fortunato2010}. This leads to an optimization problem that can be solved by performing the eigendecomposition of the Laplacian matrix  \cite{Belkin2003}. However, this technique may not scale well when the number of graph vertices grows big. 

A possible way to reduce the computational requirements of eigendecomposition involves the use of Nystr\"om method, a random sampling technique to approximate the eigenvectors of a large matrix \cite{Fowlkes2004}. Similarly, one can constrain the original eigendecomposition problem to satisfy the Nystr\"om formula, so as to obtain a reduced-size problem that achieves a lower approximation error \cite{vladymyrov16}. 

An alternative approach to speed up the computation of eigenvectors is to use the power method from the numerical linear algebra literature \cite{Boutsidis2015}. From an optimization viewpoint, the power method simultaneously updates all the coordinates of the iterate, but they converge to the optimal value at different speeds. Hence, it is possible to derive a coordinate-wise version of the power method for a further speedup \cite{Lei2016}. 

Yet another direction to efficiently deal with spectral embedding involves the use of stochastic gradient optimization. The idea is to cast the algebraic problem of identifying the principal eigenvectors of a matrix as the optimization of a trace function subject to an orthogonality constraint \cite{Xu2000, Wang2010}. Therefore, a natural way to tackle this problem consists of using the notion of Riemannian gradients \cite{Han17}.

\paragraph*{Contribution}
In this work, we propose a stochastic optimization algorithm to deal with spectral embedding. Our main contribution consists of reformulating the latter so that the orthogonality constraint is replaced with an orthogonalization matrix in the criterion. As the gradient can be computed through a Cholesky factorization, our reformulation enables us to develop an efficient algorithm based on mini-batch gradient descent \cite{Reddi2018, Dudar2017}. The peculiarity of our approach is that the graph nodes are processed by splitting them in chunks of manageable size. This allows us to deal with much larger graphs than those tackled by similar methods \cite{Belkin2003, Han17}. 

\paragraph*{Outline}
The paper is organized as follows. In Section~\ref{sec:aggregation}, we formulate the spectral embedding problem. In Section~\ref{sec:embedding}, we describe the proposed reformulation and the stochastic gradient optimization algorithm to solve it. In Section~\ref{sec:results}, we compare our approach to similar spectral embedding methods. Finally, the conclusion is drawn in Section~\ref{sec:conclusion}.

\section{Problem formulation}
\label{sec:aggregation}

In this paper, we aim at clustering a graph $\mathcal{G}(V,E,W)$ which consists of a set $V$ of vertices, and a set $E$ of edges with non-negative weights $w_{i,j}$, where $W = [w_{i,j}]_{i,j} \in \mathbb{R}^{N\times N}$ is the weighted adjacency matrix, with $w_{i,i} = 0$ for all $i$. 

The degree of a vertex $i$ in the graph $\mathcal{G}$, denoted as $d(i)$, is defined as the sum of weights of all the edges incident to $i$ in the graph $\mathcal{G}$. The degree matrix $D$ is then defined as
\begin{equation}
D_{i,j} = \begin{cases} d(i) \quad & \textup{if $i=j$} \\
0 \quad & \textup{otherwise.}
\end{cases}
\end{equation}
Based on $W$ and $D$, the Laplacian matrix of $\mathcal{G}$ is
\begin{equation}
L = D - W.
\end{equation}
In the literature, $L$ is called the unormalized Laplacian matrix. There are also normalized versions of this matrix, such as
\begin{equation}
L_{\rm{sym}} = D^{-\frac{1}{2}}(D-W)D^{-\frac{1}{2}}.
\end{equation}

The Laplacian matrix is of broad interest in the studies of spectral graph theory \cite{Chung1997}. Although closely related to each other, the different definitions of the Laplacian matrix don't enjoy the same properties. For example, $L$ and $L_{\rm{sym}}$ are both real and symmetric matrices, but they  have a different set of eigenvalues and orthonormal eigenvectors. Moreover, each definition of the Laplacian matrix comes with its own algorithm to perform spectral embedding and clustering \cite{Luxburg2007}. In this paper, we use the unnormalized graph Laplacian $L$ defined above, since its eigenvectors are orthonormal and its eigenvalues are non-negative, with the smallest equal to zero.

\subsection{Spectral Clustering}
\label{subsec:single_layer_clustering}

Spectral clustering is a standard application of spectral embedding \cite{Malik2000, Ng2001,Luxburg2007}. The idea is to represent the data points to be clustered as the vertices of a graph, whose edges describe the pairwise relationships between those points (e.g., computed with k-nearest neighbours).
The core operation is to embed the graph vertices in a low-dimensional space, where the projected points can be trivially clustered. More specifically, for a $K$-way clustering, one aims at finding the semi-orthogonal matrix $U = \left[ u_1 \,|\dots|\, u_N \right]^\top$ such that its rows $u_n \in \mathbb{R}^K$ minimize the pairwise distances on the graph edges, namely
\begin{align}\label{eq:spectral_embbedding}
\operatorname*{minimize}_{U\in\mathbb{R}^{N\times K}}\;\;  \operatorname{Tr}(U^\top LU) := \sum_{i=1}^N\sum_{j=1}^N w_{i,j} \|u_i - u_j\|^2 \nonumber  \\
\quad{\rm s.t.}\quad U^\top U = {\bf I_{K\times K}}.
\end{align}
Each row of the matrix $U$ is a point in the low-dimensional space $\mathbb{R}^K$ representing a graph vertex. Strongly-connected vertices are thus mapped to close vectors in $\mathbb{R}^K$. One can apply a $K$-mean algorithm on the rows of the solution $\bar{U}$, so that $K$ clusters are formed by grouping together the vertices that are the most strongly connected by the graph.

\section{Proposed algorithm}
\label{sec:embedding}

It is well known that the solution to Problem \eqref{eq:spectral_embbedding}  is the matrix formed by the eigenvectors associated to the  $K$ smallest eigenvalues of the Laplacian matrix $L$. While there exist many efficient algorithms for spectral embedding \cite{Belkin2003, Fowlkes2004, Boutsidis2015, vladymyrov16, Lei2016, Han17}, they may not always scale well as the graph size $N$ grows big. Therefore, we propose an alternative approach to solve Problem \eqref{eq:spectral_embbedding}, which consists of reformulating the latter so that it can be dealt with a stochastic optimization algorithm, such as mini-batch gradient descent \cite{Reddi2018, Dudar2017}.

\subsection{Forward-backward splitting}
The main difficulty in solving Problem \eqref{eq:spectral_embbedding} via gradient descent arises from the orthogonality constraint. A possible way to circumvent this issue is to make use of forward-backward splitting \cite{Combettes2005}, which boils down to the following iterations:
\begin{equation}
(\forall t\in\mathbb{N})\qquad U_{t+1} = {\sf QR}\Big( U_t - \gamma_t LU_t \Big).
\end{equation}
Hereabove, $LU$ is the gradient of $\frac{1}{2}{\rm Tr}(U^\top LU)$, whereas the operator ${\sf QR}(\cdot)$ extracts the semi-orthogonal matrix $Q$ obtained with the QR decomposition of $U_t - \gamma_t LU_t$. Indeed, any rectangular matrix $U\in\mathbb{R}^{N\times K}$ with $N\ge K$ can be factorized as $U=Q R$, where $Q\in\mathbb{R}^{N\times K}$ is semi-orthogonal, and $R\in\mathbb{R}^{K\times K}$ is upper triangular. This was the approach followed in \cite{Han17}, where a Riemannian gradient was used instead of the regular gradient.

\subsection{Reformulation with implicit constraint}
We propose an alternative approach to deal with the orthogonality constraint in Problem \eqref{eq:spectral_embbedding}. Indeed, the latter can be enforced implicitly by using the upper triangular matrix $R$ of the QR decomposition of $U$, which is defined as
\begin{equation}
U = Q R.
\end{equation}
When $U$ is positive definite, the QR decomposition is unique, and $R$ is equal to the upper triangular factor of the Cholesky decomposition  $U^\top U =  R^\top R$. Then, it is possible to extract the Q-factor of $U$ as 
\begin{equation}
Q = U R^{-1}.
\end{equation}
This equality allows us to reformulate Problem \eqref{eq:spectral_embbedding} as
\begin{equation}  
\label{eq:embedding_reformulated}
\operatorname*{minimize}_{U\in\mathbb{R}^{N\times K}}\;\; J(U):={\rm Tr}\Big((U R^{-1})^\top L (U R^{-1})\Big).
\end{equation} 
Note that the matrix $U$ is not semi-orthogonal, but $U R^{-1}$ is. Also, $R$ is a function of $U$, for which we can compute the gradient w.r.t.\ $U$ \cite{Murray2016}. Therefore, the reformulated problem can be solved using standard gradient descent, yielding
\begin{equation}\label{eq:gradient_descent}
(\forall t\in\mathbb{N})\qquad U_{t+1} = U_t - \gamma_t \nabla J(U_t).
\end{equation}

\subsection{Stochastic gradient descent}\label{sec:sgd}
The main advantage in solving Problem \eqref{eq:embedding_reformulated} is that we can modify the update in Eq. \eqref{eq:gradient_descent} by replacing the function $J$ with a stochastic approximation $J_t$, yielding
\begin{equation}\label{eq:stochastic_gradient_descent}
(\forall t\in\mathbb{N})\qquad U_{t+1} = U_t - \gamma_t \nabla J_t(U_t).
\end{equation}

Before we can define the above function $J_t$, let us denote by $\mathcal{I}$ the set of indexes associated to the nonzero elements in the upper triangular part of Laplacian matrix $L$, namely
\begin{equation}
\mathcal{I} =\Big\{(i,j)\in\{1,\dots,N\}^2 \;|\; w_{i,j} \neq 0 \;{\rm and}\; j > i\Big\}.
\end{equation}
The Laplacian matrix can be thus decomposed as follows
\begin{equation}
L = \sum_{(i,j)\in\mathcal{I}} \underbrace{ w_{i,j} \big(e_i e_i^\top + e_j e_j^\top - e_i e_j^\top - e_j e_i^\top\big) }_{\mathsf{L}_{i,j} \in \mathbb{R}^{N\times N}},
\end{equation}
where $e_n$ is the $n$-th column of the $N\times N$ identity matrix. Thus, the original criterion in Problem \eqref{eq:spectral_embbedding} can be rewritten as
\begin{equation}
\operatorname{Tr}(U^\top LU)  = \sum_{(i,j)\in\mathcal{I}} \operatorname{Tr}(U^\top \mathsf{L}_{i,j} U) = \sum_{(i,j)\in\mathcal{I}} w_{i,j} \|u_i - u_j\|^2.
\end{equation}

To build the function $J_t$ appearing in Eq. \eqref{eq:stochastic_gradient_descent}, at every iteration $t$ we select a subset of indexes
\begin{equation}
\mathcal{S}_t \subset \mathcal{I},
\end{equation}
we use those indexes to build a partial Laplacian matrix
\begin{equation}
\mathbf{L}_t = \sum_{(i,j)\in\mathcal{S}_t} \mathsf{L}_{i,j},
\end{equation}
and we approximate the criterion in Eq. \eqref{eq:embedding_reformulated} as
\begin{equation}
J_t(U) := {\rm Tr}\Big((U R^{-1})^\top \mathbf{L}_t (U R^{-1})\Big).
\end{equation}
Here is where the proposed reformulation comes into play. As the term $\mathbf{L}_t$ contains many rows/columns filled with zeros, we never have to work with the full-size matrix $U$. Specifically, the function $J_t(U)$ boils down to
\begin{equation}
J_t(U) := \sum_{(i,j)\in\mathcal{S}_t} w_{i,j} \|(u_i-u_j)^\top R^{-1} \|^2.
\end{equation}
As for the Cholesky decomposition $U^\top U =  R^\top R$, we compute the full-size matrix product at the algorithm initialization
\begin{equation}
M_0 = U_0^\top U_0.
\end{equation}
Since the update in Eq.  \eqref{eq:stochastic_gradient_descent} only modifies the terms $u_i$ and $u_j$ such that $(i,j)\in\mathcal{S}_t$, we can efficiently update it as
\begin{align}
M_{t+1} = M_t &+ \sum_{(i,j)\in\mathcal{S}_t} \big(u_{i,t+1} u_{i,t+1}^\top - u_{i,t} u_{i,t}^\top \big)\nonumber\\
&+ \sum_{(i,j)\in\mathcal{S}_t} \big(u_{j,t+1} u_{j,t+1}^\top - u_{j,t} u_{j,t}^\top\big).
\end{align}
Putting all together, we arrive at the proposed optimization algorithm, which can be readily implemented with modern tools for automatic differentiation \cite{paszke2017}.

\section{Numerical results}
\label{sec:results}

We evaluate the proposed algorithm in the context of multilayer graph clustering. An overview is given in the following.

\subsection{Multilayer Graph Clustering} 
With the increasing richness of datasets, applications even rely often on multiple sources of information to characterize the relationships between data points. This leads to multi-layer graph representations, where data points are modelled as nodes shared across the layers, each of which describes a set of relationships (graph edges) on their set of observations (shared graph nodes). For example, a social network can be represented by a multilayer graph, where each layer corresponds to a different type of relationship among the same group of persons, such as friendships, music interests, etc.

More precisely, a multilayer graph consists of $S$ edge layers sharing the same vertices. Each layer $s$ is represented by a Laplacian matrix $L^s$, along with a spectral embedding matrix $U^s$ obtained as the solution to Problem \eqref{eq:spectral_embbedding}. These matrices $(L^s,U^s)_{1\le s\le S}$ can be merged into a representative embedding matrix $U$ via the minimization of the projection distance in a Grassman manifold,\footnote{By definition, a Grassmann manifold is the set of $K$-dimensional linear subspaces in $\mathbb{R}^N$. Each point in this manifold can be represented by a semi-orthogonal matrix $U \in \mathbb{R}^{N\times K}$. The distance between two points $U_1$ and $U_2$ is then defined based on a set of principal angles $\{\theta_k\}_{1\le k\le K}$ between the corresponding subspaces. These angles are the fundamental measure used to define various distances on the Grassmann manifold, such as the squared projection distance \cite{Hamm2008}, defined as
	\begin{equation*}
	d^2_{\sf proj}(U_1,U_2) 
	= K - \operatorname{Tr}(U_1U_1^\top U_2U_2^\top) \\
	\end{equation*}
}
 defined as
\begin{equation}
d_{\sf proj}(U,\{U^s\}_{1\le s\le S}) = KS - \sum_{s=1}^S \operatorname{Tr}\big(UU^\top U^s{U^s}^\top\big).
\end{equation}
Consequently, multilayer graph clustering boils down to \cite{Dong_TSP_2014}
\begin{align}\label{eq:multilayer_embedding}
\operatorname*{minimize}_{U\in\mathbb{R}^{N\times K}}\;  \operatorname{Tr}\Big( \sum_{s=1}^{S}U^\top\big( L^s - & \alpha \, U^s {U^s}^\top\big) U\Big) \nonumber  \\
& \quad{\rm s.t.}\quad U^\top U = {\bf I_{K\times K}}.
\end{align}
This is equivalent to perform spectral embedding on the aggregated Laplacian matrix defined as
\begin{equation}\label{eq:aggregated_laplacian}
L_{\sf MLG} = \sum_{s=1}^{S} (L^s - \alpha \, U^s {U^s}^\top),
\end{equation}
where $\alpha>0$ is a regularization parameter.

For the sake of completeness, note that there exist other approaches to deal with a multilayer graph, such as averaging the Laplacian matrices of individual layers \cite{argyriou_2005_nips,Tang_20012, Chen_Hero_2017}, or treating them as points in a Grassmann manifold \cite{Sindhwani2005, Kumar2011, Wang_TIP_2013, Dong_TSP_2014}.

\subsection{Datasets}
In our experiments, we consider a synthetic dataset, as well as a real dataset coming from the Yelp challenge. Both have a multilayer graph representation of data. 

As for the synthetic data, we consider a multilayer graph with $S = 3$ layers and $K=5$ clusters. Data points are drawn from a Gaussian mixture model with five components, each representing a cluster. The goal with this dataset is to recover the five clusters of the graph vertices, using the layers constructed from the point clouds.

\begin{figure}[t]
	\centering
	\includegraphics[width=\linewidth]{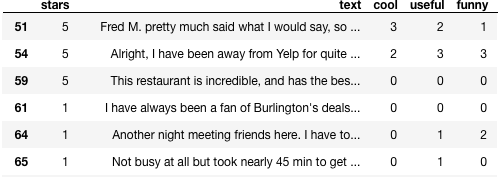}
	\caption{Yelp data before the preprocessing of text reviews.}
	\label{fig:yelp}
\end{figure}

Yelp is a popular website for reviewing and rating local businesses. In our experiments, we only extract star ratings, text reviews, and review evaluations (users can mark reviews as ``cool'', ``useful'', and ``funny''), ignoring the other information in the dataset. An example is shown in Fig. \ref{fig:yelp}. Our goal is to cluster the businesses by predicting their ratings ($K=3$).

To build a multilayer graph on Yelp data, we proceed as follows. We preprocess the text reviews using sentiment analysis. This yields a polarity score within the range $[-1, 1]$ on which we build a 5-nearest neighbor (NN) graph. We also build a 5-NN graph on the other features (``cool'', ``useful'', and ``funny'' evaluations), leading to $S=4$ graph layers.

\subsection{Comparisons}
We compare spectral embedding algorithms on Problem \eqref{eq:multilayer_embedding}, assuming that the aggregated Laplacian matrix defined in Eq. \eqref{eq:aggregated_laplacian} is given as input. This means that a spectral embedding problem is solved to compute each matrix $U^s$. For brevity, these problems are not considered into our  analysis.

\paragraph*{Problem setting}
We evaluate the performance of three spectral embedding algorithms: matrix eigendecomposition \cite{Belkin2003}, stochastic forward-backward splitting \cite{Han17}, and our method described in Section \ref{sec:sgd}. Note that ``matrix eigendecomposition`` uses the normalized version of the aggregated Laplacian matrix. The normalization is important to obtain a good clustering performance. However, this causes the method to slow down considerably, which gives rise to the interest of considering alternative approaches. The other methods use the unnormalized Laplacian matrix.

\paragraph*{Implementation details}
All the experiments were conducted in Python/Numpy on a 40-core Intel Xeon CPU at 2.5 GHz with 128GB of RAM. As for the proposed method, we used a step-size $\gamma=10^{-3}$, and mini-batches of size $4'000$. The latter is a critical parameter, as mini-batches should be big enough to capture the structure of the Laplacian matrix. For this reason, it is also highly important that mini-batches are sampled at random from the entire dataset at each step, and not be fixed across iterations.

\paragraph*{Result analysis}
We use three criteria to measure the clustering performance: Purity, Normalized Mutual Information (NMI), and Rand Index (RI). The results reported in Tables \ref{tab:synt} and \ref{tab:real} for both synthetic and Yelp data show that the proposed method achieves almost the same clustering performance as the matrix eigendecomposition, while being faster than the other compared methods. The small differences in the performance are related to the stochastic nature of our optimization algorithm. Running it for less iterations would allow us to get even faster results, yet less accurate. There is thus a tradeoff between speed and accuracy, as usual in optimization.

\begin{table}[t]
\centering
\caption{Clustering on synthetic data ($N=10'000$). \label{tab:synt}}
\begin{tabular}{ l   c  cc   c   c c}
\toprule
Method & Time & Iter.& Purity & NMI & RI \\
\midrule	 
Eigendecomp. & 36.3 s & - & 0.93 & 0.83 & 0.85 \\
Stochastic FB & 22.8 s & 3000 & 0.94 & 0.84 & 0.86 \\
Proposed & 5.2 s & 500 & 0.92 & 0.79 & 0.82 \\
\bottomrule	
\end{tabular}
\end{table}

\begin{table}[t]
\caption{Clustering on Yelp data ($N=11'160$). \label{tab:real}}
\centering
\begin{tabular}{ l   cc   c   c   c c}
\toprule
Method & Time & Iter. & Purity & NMI & RI \\
\midrule	 
Eigendecomp. & 69.7 s & - & 0.87 & 0.65 & 0.56\\
Stochastic FB & 50.6 s & 3000 & 0.88 & 0.79 & 0.88 \\
Proposed & 11.4 s & 500 & 0.86 & 0.71 & 0.84 \\
\bottomrule	
\end{tabular}
\end{table}

\section{Conclusion}
\label{sec:conclusion}

In this paper, we proposed a reformulation of spectral embedding that allowed us to devise a new optimization algorithm based on stochastic gradient descent. The experimental results showed that the proposed approach is more efficient than matrix eigendecomposition and stochastic Riemannian optimization on medium-size datasets. This confirms that our approach is capable of scaling to larger datasets than what similar spectral embedding methods can handle.

There are a number of extensions that we are currently investigating, such as the possibility to extend our approach to the Riemann manifold of semi-positive definite matrices \cite{ElGheche2018_arxiv}, and to simultaneously compute the embedding matrices of each layer and the aggregated graph. Moreover, a GPU implementation of the proposed algorithm is under development.

\bibliographystyle{IEEEbib}
\bibliography{strings,biblio}

\end{document}